\documentclass{article}

\usepackage{arxiv}

\usepackage[utf8]{inputenc} % allow utf-8 input
\usepackage[T1]{fontenc}    % use 8-bit T1 fonts
\usepackage{hyperref}       % hyperlinks
\usepackage{url}            % simple URL typesetting
\usepackage{booktabs}       % professional-quality tables
\usepackage{amsfonts}       % blackboard math symbols
\usepackage{nicefrac}       % compact symbols for 1/2, etc.
\usepackage{microtype}      % microtypography
\usepackage{lipsum}
\usepackage{graphicx}
\usepackage{amsmath} 
\graphicspath{ {./images/} }

\title{Predict future sale}

\title{Comparative Study of Domain Driven Terms Extraction Using Large Language Models}

\author{
 Sandeep Chataut \\
  University of South Dakota\\
  \texttt{sandeep.chataut@coyotes.usd.edu} \\
   \And
 Tuyen Do \\
  University of South Dakota\\
  \texttt{tuyen.do@usd.edu} \\
  \And
 Bichar Dip Shrestha Gurung \\
  University of South Dakota\\
  \texttt{bichar.shresthagurun@coyotes.usd.edu} \\
    \And
 Shiva Aryal \\
  University of South Dakota\\
  \texttt{shiva.aryal@coyotes.usd.edu} \\
    \And
 Anup Khanal \\
  University of South Dakota\\
  \texttt{anup.khanal@coyotes.usd.edu} \\  
      \And
 Carol Lushbough \\
  University of South Dakota\\
  \texttt{Carol.Lushbough@usd.edu} \\  
      \And
 Etienne Gnimpieba \\
  University of South Dakota\\
  \texttt{etienne.gnimpieba@usd.edu} \\ 
}

\begin{document}
\maketitle

\begin{abstract}
Keywords play a crucial role in bridging the gap between human understanding and machine processing of textual data. They are essential to data enrichment because they form the basis for detailed annotations that provide a more insightful and in-depth view of the underlying data. Keyword/domain driven term extraction is a pivotal task in natural language processing, facilitating information retrieval, document summarization, and content categorization. This review focuses on keyword extraction methods, emphasizing the use of three major Large Language Models(LLMs): Llama2-7B, GPT-3.5, and Falcon-7B. We employed a custom Python package to interface with these LLMs, simplifying keyword extraction. Our study, utilizing the Inspec and PubMed datasets, evaluates the performance of these models. The Jaccard similarity index was used for assessment, yielding scores of 0.64 (Inspec) and 0.21 (PubMed) for GPT-3.5, 0.40 and 0.17 for Llama2-7B, and 0.23 and 0.12 for Falcon-7B. This paper underlines the role of prompt engineering in LLMs for better keyword extraction and discusses the impact of hallucination in LLMs on result evaluation. It also sheds light on the challenges in using LLMs for keyword extraction, including model complexity, resource demands, and optimization techniques.
\end{abstract}

% \begin{keywords}
% Natural Language Processing, Large Language Models, data enrichment
% \end{keywords}
% keywords can be removed
\keywords{Natural Language Processing\and Large Language Models\and data enrichment}

\section{Introduction}
The discipline of natural language processing has seen a revolution in recent years due to the swift development of Large Language Models (LLMs)\cite{Yang:23}. The capabilities of LLMs are remarkable considering the seemingly straightforward nature of the training methodology \cite{Touvron:23}. LLM has started to outperform many traditionally used machine learning models with its evolution. These models have been pre-trained on sizable text corpora, picking up complex syntactic and semantic structures and have demonstrated remarkable abilities in a multitude of tasks, ranging from tasks like text generation, sentiment analysis, tabular data embedding\cite{Do2023} named entity recognition \cite{Bomgni2023}, detection of adverse drug effects \cite{Bom2023} and many more. One such prominent application is keyword extraction, a fundamental task in information retrieval, document summarization, and content categorization. 

In this context, we have developed a Python package tailored for keyword extraction using these advanced LLMs. This tool represents an integral part of our research, facilitating the practical application of LLMs in extracting relevant keywords from varied text corpora. Keywords play a pivotal role in bridging the gap between human understanding and machine processing of textual data. Keywords are an important asset to various search engines as well as many machine learning frameworks as they act as primary features for NLP or any ML tasks \cite{Do:24}. Keywords are crucial to data enrichment because they form the basis for detailed annotations that provide a more insightful and in-depth view of the underlying data. They serve to efficiently index content, summarize documents, and retrieve information since they capture the main ideas and topics in a work. The strategic use of keywords serves as a powerful tool for enriching data, as it enables the addition of comprehensive annotations that enhance the overall context. 

Traditional methods of keyword extraction, often reliant on statistical metrics and linguistic patterns, have yielded valuable insights. However, the advent of LLMs has ushered in a new era of keyword extraction, where contextual nuances, semantic relationships, and domain-specific intricacies can be captured with unparalleled accuracy. Traditional keyword extraction methods including frequency based, graph based and statistical approaches have been foundational in the field of Natural Language Processing (NLP). Information retrieval systems frequently employ frequency-based techniques, such as Term frequency–inverse document frequency (TF-IDF), which is one of the most widely used term weighting schemes. By assigning a word a score based on the corpus to which the document belongs, TF-IDF is a numerical statistics that reveals the relative importance of a word in a text \cite{Amir:21}. Graph-based methods, such as TextRank, use graph structures to represent term relationships \cite{Tarau:04}. Modern Large Language Models (LLMs) have demonstrated enhanced capabilities in capturing context and semantics, making them more suitable for keyword extraction in complex language domains.

As the field advances, there is a growing need for comprehensive evaluations and comparative analyses to guide researchers and practitioners in choosing the most suitable LLM for specific tasks. This review paper aims to contribute to the burgeoning field of keyword extraction by offering a comprehensive analysis of the state-of-the-art large language models. We present an in-depth evaluation of their performance on the Inspec and PubMed dataset, a benchmark collection of scientific literature. Each model is examined individually to understand its unique approach to keyword extraction. This paper also highlights the innovative use of prompt engineering techniques in LLMs for enhancing keyword extraction. This study aims to contribute insights into the performance of GPT 3.5, LLama2-7B, and Falcon-7B in keyword extraction, with a focus on both quantitative and qualitative assessments. Within LLMs, hallucination is a more expansive and all-encompassing notion that mostly focuses on factual inaccuracies \cite{Huang:23}. This study also addresses the phenomenon of hallucination in LLMs, impacting the overall evaluation and interpretation of the results. Understanding and addressing these hallucination-related challenges are crucial for optimizing the performance of LLMs in domain driven term extraction tasks. Furthermore, a comparative analysis sheds light on their relative performance, highlighting the advancements they bring to the table.
\section{Methods}

\subsection{Dataset Selection}
Experiments using the Inspec and PubMed datasets, which are representative of a number of disciplines covered by scientific literature, provide insights that are used in this study. The 500 abstracts of Computer Science scientific journal papers that were gathered between 1998 and 2002 make up Inspec. There are two sets of keywords assigned to each document: the uncontrolled keywords are those that are freely assigned by the editors and are not limited to the thesaurus or the document, and the controlled keywords are those that are manually controlled assigned keywords that appear in the Inspec thesaurus but may not appear in the document \cite{Hulth:03}.

On the other hand, the PubMed dataset comprises 500 abstracts meticulously selected to delve into the intersection of biofilms and materials, reflecting the pivotal role of these subjects in contemporary biomedical research . The keywords used for comparing PubMed data are extracted from the same 500 papers using the index terms. For comparison and analysis, we consider a union of both sets as the ground-truth. This paper examines comparative evaluations of the performance of the selected large language models while evaluating them on the 500 texts of Inspec dataset and 500 abstracts from PubMed \cite{Aronson:00}.

\subsection{Evaluation Metrics}
There are several evaluation metrics to assess the quality and effectiveness of keyword extraction methods. When comparing the keywords extracted from our Large Language Models (LLMs) with the reference keywords from both Inspec dataset and PubMed datasets. The Jaccard similarity statistic was employed in our review work. A metric for comparing two sets' similarity is the Jaccard similarity, sometimes referred to as the Jaccard coefficient or Jaccard index. It can be calculated by dividing the size of the intersection of the sets by the size of their union \cite{Costa:21}.
\begin{equation}\label{Jaccard_index}
    J(A,B) = \frac{|A \cap B|}{|A \cup B|}
\end{equation}
Where, A and B are sets.

$|A \cap B|$ \text{ is the size of the intersection of sets A and B and}

$|A \cup B|$ \text{ is the size of union of sets A and B. }
In the context of keyword extraction, sets A and B represent the sets of keywords extracted by the model and the reference Inspec keywords, respectively for the first evaluation. Second evaluation is done by taking sets A and B represent the sets of keywords extracted by the model and the index terms extracted from Pubmed papers .
\subsubsection{Interpretation}
A Jaccard similarity of 1 indicates a perfect match between the model's keywords and the reference keywords while Jaccard similarity of 0 indicates no overlap between the sets, meaning that none of the model's keywords match the reference keywords.

\subsubsection{Use in Comparative Analysis}
When comparing different large language models (LLMs) such as GPT 3.5, LLama2-7B, and Falcon-7B, the Jaccard similarity can be calculated for each model's set of keywords against the reference set. This allows for a quantitative assessment of the models' performance in terms of keyword extraction accuracy.

\subsection {Framework Utilized and Package Architecture}
The implementation of this research was greatly aided by LangChain, a flexible framework for creating applications powered by language models. An open-source framework called LangChain was created to make it easier to create applications that use LLMs. It provides an array of parts, tools, and interfaces that make building LLM-focused applications easier. LangChain is an excellent tool for integrating language models with a variety of context sources, such as prompt instructions, external material, and few-shot examples. It was designed to give programs contextual awareness \cite{Langchain:22}.

In conjunction with our investigation into LLMs for keyword extraction, we developed a specialized Python package. This package is seamlessly integrated with the LangChain framework, facilitating efficient interaction with various LLMs such as Llama-2 7B, GPT-3.5, and Falcon-7B. Central to this package is the LLM\_KeywordExtractor class, which encapsulates the functionality for initializing different LLMs and extracting keywords based on customized prompt engineering techniques. The package is designed to offer flexibility, allowing for the easy addition of new models or modification of existing prompt templates.

\subsection {Keyword Visualization with Word Clouds}
We also used word clouds to visually represent the most frequently occurring words in the keywords matching process. In word clouds, words are displayed in varying sizes based on their frequency. 
\begin{figure*}[t]
\centering
\includegraphics[width=160mm,height = 5cm]{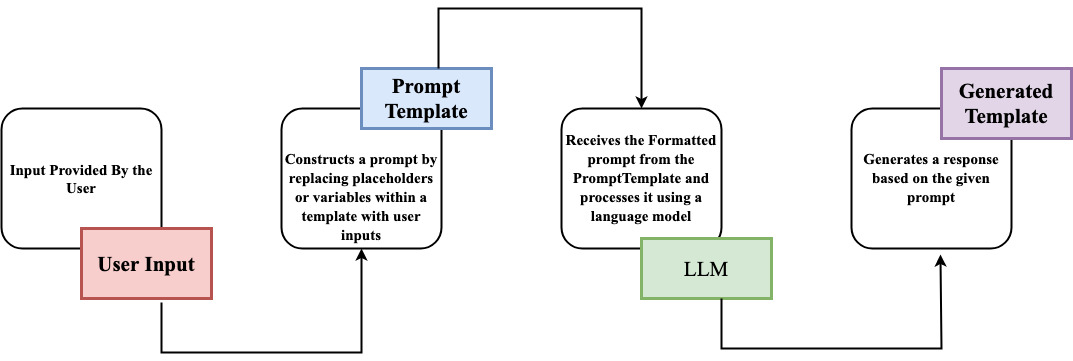}
\caption{Workflow of Langchain framework}
\label{Langchain}
\end{figure*}

\subsection{Prompt Engineering Techniques}

In the field of LLMs, the practice of prompt engineering is vital for directing these models to efficiently execute specific tasks. Prompt engineering, as a new field, is concerned with creating the best instructions for LLMs to accomplish tasks successfully (\cite{Giglou:23}). This paper provides a concise overview of various prompt engineering methods. Additionally, we describe the specific technique employed in our research. Some of the prompt engineering techniques are discussed below: 

Chain-of-Thought (CoT) Prompting: This method involves crafting prompts that guide LLMs through a logical, step-by-step reasoning process to solve complex tasks. It is particularly useful in breaking down intricate tasks into simpler, incremental steps \cite{Giglou:23}.

Tree-of-Thoughts (ToT) Prompting: An extension of CoT, ToT prompting is used for tasks requiring exploration or strategic lookahead. It aids LLMs in exploring intermediate steps for general problem-solving, enhancing their capability to handle complex reasoning tasks \cite{Giglou:23}.

Cloze and Prefix Prompts: These involve creating prompts where LLMs either fill in the blanks (cloze prompts) or generate text following a given starting phrase (prefix prompts). Each type offers different adaptabilities and is chosen based on the task's requirements and the model's pretraining objectives \cite{Giglou:23}.

Customized Prompt Design: This approach involves crafting unique prompts tailored to the specific requirements of a task, allowing for a more focused and effective interaction with LLMs \cite{Giglou:23}.

In this study, we have employed a customized zero shot prompting that aligns with the goals of our task. We use a standardized system prompt that instructs the LLM to function as an expert in keyword extraction. This prompt is designed to be broad yet specific enough to guide the LLM towards identifying main ideas, concepts, entities, or themes in the input text.

We introduce a function $f_{\text{KeywordExtraction}}(P, L)$ that formalizes the construction of our prompts, as shown in Equation \ref{Prompt}. This function takes two parameters: 

P, which encapsulates the attributes of the assistant, and 

L, which is the input text upon which keyword extraction is to be performed. The output of this function is a structured prompt that combines these elements and introduces a [MASK] placeholder, indicating where the model's output, i.e., the extracted keywords, will be populated.

Equation \ref{Prompt2} defines the interaction with the LLMs, where the output response is generated as a result of the LLM processing the prompt constructed by $f_{\text{KeywordExtraction}}$.

The output response is the LLM's interpretation and execution of the task, resulting in a set of keywords that best represent the main ideas, concepts, entities, or themes within the provided text.

To visualize this process, the flow diagram (refer to Figure 1) elucidates each stage of interaction. The User Input corresponds to L, the input text. The Prompt Template correlates with the attributes P and the application of $f_{\text{KeywordExtraction}}$ to create the prompt. The LLM represents the computational process where the LLM interprets and processes the prompt. Finally, the Generated Template is reflective of the [MASK] being filled with the output response, which is the set of extracted keywords.

\begin{equation}\label{Prompt}
f_{\text{KeywordExtraction}}(P, L) = \text{[}P\text{].[}L\text{] [MASK]} 
\end{equation}

\begin{equation}\label{Prompt2}
    \text{Output~response} = \text{LLM}(f_{\text{KeywordExtraction}}(P, L))
\end{equation}

where,
$f_{\text{KeywordExtraction}}(P, L)$ is the function that constructs the prompt for keyword extraction.

P denotes the attributes of the assistant, which in our context, include being helpful, respectful, honest, and an expert in keyword extraction. These attributes are articulated to prime the LLM for the task, facilitating a more focused and accurate extraction of keywords that represent the main ideas, concepts, entities, or themes in the input text.

L is the input text upon which the keyword extraction is to be performed.

[MASK] serves as a placeholder within the prompt, indicating where the LLM's output, the extracted keywords, is expected to be inserted.

LLM refers to the specific Large Language Model employed for the keyword extraction task and output response is the set of keywords generated by the LLM as a response to the prompt, encapsulating the essence of the input text

% \begin{equation}
% \begin{aligned}
% f_{\text{KeywordExtraction}}(\text{input\_text}) = \text{[system\_prompt]} + \text{input\_text}
% \end{aligned}
% \end{equation}
% where, \\

% $f_{\text{KeywordExtraction}}(\text{input\_text})$ is the function for the keyword extraction prompt.

% "[system\_prompt]" represents the system prompt which is instruction to the model: "You are a helpful, respectful, and honest assistant. You are an expert in keyword extraction. You are provided texts from various contexts. Your task is to generate a set of relevant keywords from the input text provided. The keywords should ideally represent the main ideas, concepts, entities, or themes mentioned in the input text."

% and \text{"input\_text"} is the placeholder for the input text to be analyzed.

\section{Large Language Models}
LLMs exhibit strong capacities to understand natural language and solve complex tasks (via text generation)\cite{Zhao:23}. As we delve into the capabilities and architectures of Large Language Models, it becomes crucial to understand the underlying mechanisms that enable these models to perform tasks like keyword extraction with remarkable efficiency. A cornerstone of these models, particularly relevant to the Transformer architecture, is the attention mechanism. The scaled dot-product attention, a key component of this mechanism, allows the model to dynamically weigh the significance of different parts of the input data\cite{Vaswani:23}. Mathematically, it is represented as:

% \text{Attention}(Q, K, V) = \text{softmax}\left(\frac{QK^T}{\sqrt{d_k}}\right)V

\begin{equation}
\text{Attention}(Q, K, V) = \text{softmax}\left(\frac{QK^T}{\sqrt{d_k}}\right)V
\end{equation}

In this equation, \(Q\), \(K\), and \(V\) stand for the queries, keys, and values matrices, respectively, which are derived from the input embeddings. The term \(d_k\) represents the dimensionality of the keys, and the softmax function is applied to the result of the matrix multiplication of \(Q\) and \(K^T\), scaled by \(d_k\), ensuring that the weights sum to 1. This operation is followed by the multiplication of the softmax output with the values matrix \(V\), culminating in an output that reflects the aggregated attention across the input data.

In the realm of natural language processing, particularly in tasks such as keyword extraction, the order of words in a sentence carries significant meaning. Transformer models, which are at the heart of many LLMs, do not inherently account for the order of input data they treat input sequences as sets rather than ordered lists. To rectify this and enable the model to leverage the sequential nature of language, positional encoding is used.
Positional encoding vectors are added to the embedding vectors to provide position information \cite{Vaswani:23}. The model can then use these vectors to determine where each word is in the sentence. This is crucial for understanding the context and semantics necessary for tasks like keyword extraction, where the relevance of a term can be position-dependent.

\begin{equation}
PE(pos, 2i) = \sin\left(\frac{pos}{10000^{\frac{2i}{d_{\text{model}}}}}\right)
\end{equation}

\begin{equation}
PE(pos, 2i+1) = \cos\left(\frac{pos}{10000^{\frac{2i}{d_{\text{model}}}}}\right)
\end{equation}

where \(PE\) is the positional encoding, \(pos\) is the position in the sequence, \(i\) is the dimension, and \(d_{\text{model}}\) is the dimension of the model's embeddings. The use of sine and cosine functions helps the model to easily determine the position of each word in the sequence. By applying these functions to even and odd positions within the encoding vector, the model retains a unique positional signature for each word.

The subject of natural language processing has advanced significantly in recent years, mostly due to the creation of large language models (LLMs) that demonstrate an unparalleled comprehension of human language. This review paper primarily focuses on the variety of keyword extraction methods, concentrating on the use of three significant LLMs: Llama-2 7B, GPT-3.5 and Falcon-7B.

\subsection{Llama2-7B}
Llama2-7B, an improved iteration of Llama 1, was trained using a fresh set of openly accessible data. In addition, grouped-query attention is used, the pretraining corpus size is doubled, and the context length of the model is raised by 40\%. Rejection sampling and proximal policy optimization (PPO), two techniques used in Reinforcement Learning with Human Feedback (RLHF) techniques, are used to iteratively improve the model \cite{Touvron:23}. The version of Llama2 we are using in this paper is pretrained and fine-tuned generative text model ranging having 7 billion parameters.

\subsection{GPT-3.5}
 New developments in the field of natural language processing, including the 175 billion parameters of GPT-3, have fueled the development of the GPT-3.5 model. With fewer parameters, GPT-3.5 is an improved version of GPT-3 that incorporates machine learning algorithm fine-tuning. The fine-tuning process involves reinforcement learning with human feedback, which helps to improve the accuracy and effectiveness of the algorithms \cite{Chen:23}.
\subsection{Falcon-7B}
Falcon-7B is a 7B parameters causal decoder-only model built by TII. It was trained on 1,500B tokens of Refined Web, a high-quality filtered and deduplicated web dataset which is enhanced with curated corpora. It is made available under the Apache 2.0 license \cite{Penedo:23}.
\section{Results and Discussion}
\subsection{Quantitative Results}
\subsubsection{Jaccard Similarity Scores}
\begin{figure}[htbp]
\centerline{\includegraphics[width=90mm,height=5.5cm]{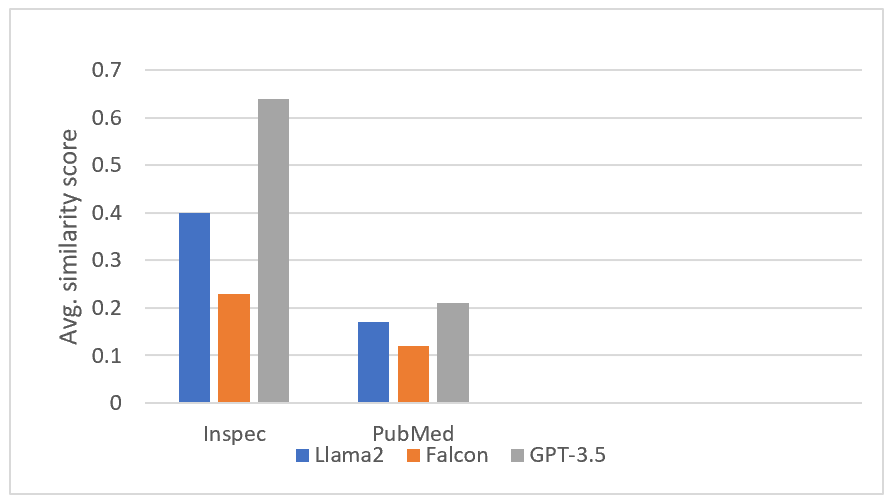}}
\caption{Comaprison of Jaccard Similarity score}
\label{Jaccard}
\end{figure}

The average Jaccard similarity score of 500 texts from Inspec dataset and 500 abstracts from PubMed dataset calculated for each LLMs is shown in Figure \ref{Jaccard}.

\subsubsection{Observations}
From figure \ref{Jaccard}, we can see that GPT-3.5 demonstrated the highest average Jaccard similarity score of 0.64 and 0.21 for Inspec and PubMed datasets respectively, indicating a substantial overlap between its generated keywords and the reference Inspec and PubMed keywords. LLama2-7B achieved score of 0.40 while comparing with Inspec dataset and 0.17 with PubMed dataset demonstrating their lower effectiveness while calculating similarity index. Falcon-7B showed a very lower degree of overlap with the reference keywords with a score of 0.23 with Inspec dataset and 0.12 with PubMed dataset.

In the process of extracting domain specific terms from the models, a specific temperature parameter was employed and the value was set to 0.2. The models' behavior during keyword extraction is greatly influenced by this temperature selection. One parameter that regulates the output of the LLM's randomness is the temperature. A higher temperature will result in more creative and imaginative text, while a lower temperature will result in more accurate and factual text. In our case, lower temperature value of 0.2, tend to yield more predictable and narrowly focused results, which may have an effect on the variety and originality of the generated keywords. Because it affects the balance between precision and diversity in the model outputs, the effect of this temperature setting must be taken into account when evaluating and interpreting the keyword extraction findings.

\subsubsection{Package Performance and Utility}
Our package's ability to interface with multiple LLMs and efficiently extract keywords was instrumental in our research. It demonstrated not only a high degree of accuracy, as reflected in the Jaccard similarity scores, but also offered ease of use and adaptability to different text corpora. 

\subsubsection{Analysis of Llama2-7B}

\paragraph{Additional Keywords}

Llama2-7B exhibits cases where it generates additional keywords that are not present in the reference Inspec and PubMed sets.  This suggests that Llama2-7B has the capacity to identify terms that may be overlooked in the reference sets. Despite generating additional keywords, Llama2-7B maintains a degree of similarity with reference sets in some cases. However, the overall Jaccard similarity scores for PubMed sets may be lower due to the introduction of new terms and also because the reference PubMed index terms were very less in comparison to keywords extracted from Llama2-7B.
\paragraph{Keywords with Definitions}

In some cases, Llama2-7B extracted keywords along with brief definitions enhancing clarity. While this provides valuable insights, it contributes to a larger dissimilarity in Jaccard similarity scores due to the introduction of novel terms.

These observations highlight the impact of hallucinations on Llama2-7B's keyword extraction, emphasizing the need for a careful examination of the trade-offs between introducing new, potentially relevant terms and maintaining similarity with reference sets
\subsubsection{Analysis of GPT-3.5}

GPT-3.5 demonstrates a commendable capability in generating keywords that closely match the manually generated reference set. This alignment shows how well the model captures the main ideas of the content and how accurate it is. The absence of unnecessary information in GPT-3.5’s output contributes to user-friendliness and the focus on generating key words only contributes to the model’s clarity and efficiency. However, the overall Jaccard similarity scores for PubMed sets may be lower due to the introduction of few new terms that may be associated with hallucinations and also because the reference PubMed index terms were very less in comparison to keywords extracted from GPT 3.5.
\subsubsection{Analysis of Falcon-7B}

Falcon-7B exhibits a relatively low Jaccard similarity score, indicating a limited overlap between the keywords generated and the generated reference set. This suggests a lower degree of concordance in the identified terms compared to other models. It is found that unnecessary words were also extracted along with very few relevant keywords and has impacted the overall quality of the extracted keywords. Moreover, the finding that relatively few relevant keywords are retrieved suggests a behavior in Falcon-7B's output that is similar to hallucination.
\subsubsection{Analysis of factors affecting keyword extraction performance}

We noted the proficiency of the Llama2-7B, GPT3.5, and Falcon-7B models in several areas when we examined their keyword extraction ability. Strong contextual comprehension exhibited by the models influences careful keyword selection. The models' ability to closely align with reference terms is impacted by domain-specific problems, such as evolving language use or subtle terminology, which in turn affects similarity scores. Variability in the training datasets for each model results in differences in their generalization to specific keyword extraction tasks.
Models struggle with domain-specific terminology and context if the training data does not fully cover the texts in the keyword extraction task, which results in subpar performance. It's critical to keep the delicate balance between task relevance and model complexity. For certain keyword extraction jobs, models need to be adjusted to better fit the specifics of the target domain. A mismatch between task needs and model expectations could result from improper fine-tuning. Due to the unique linguistic and contextual factors involved in keyword extraction tasks, models must have strong mechanisms in place to adapt and reliably extract information.
\subsection{Qualitative Analysis}

\subsubsection{Inference Time}
Inference time, often referred to as latency or response time, is a critical performance metric when assessing the suitability of Large Language Models (LLMs) for various applications. It measures the time taken by an LLM to process and generate results in response to a given input query. It is crucial to comprehend the inference time of various LLMs when it comes to keyword extraction, particularly for applications that require speed or real-time processing. The inference time and hardware specifications for each language model, are summarized in the Table I:

% \begin{table}[h]
%     \centering
%     \caption{Inference time comparison}
%     \small % Adjust font size if necessary
%     \begin{tabular}{| p{1.8cm} | p{2.6cm} | p{3.5cm} | p{3.cm} |} 
%     \hline 
%     \textbf{LLM} & \textbf{Avg Inference Time} & \textbf{Hardware Specifications} & \textbf{Remarks} \\ [1ex] 
%     \hline 
%     Falcon 7B & 7-12 secs & T4 GPU & Small variation in inference time based on different input lengths \\ [1ex]
%      \hline 
%     Llama2 7B & 4-8 secs & T4 GPU & Small variation in inference time based on different input lengths \\ [1ex]
%     \hline 
%     GPT 3.5 & 3-5 secs & CPU & Almost no variation in inference time based on different input lengths \\ [1ex]
%     \hline  
% \end{tabular}
% \end{table}

\begin{table}[h]
    \centering
    \small % Adjust font size if necessary
    \begin{tabular}{|p{1.8cm}|p{2.6cm}|p{3.5cm}|p{4.5cm}|} 
    
    \textbf{LLM} & \textbf{Avg Inference Time} & \textbf{Hardware Specifications} & \textbf{Remarks} \\ \hline
    Falcon-7B & 7-12 secs & T4 GPU & Small variation in inference time based on different input lengths \\
    Llama2-7B & 4-8 secs & T4 GPU & Small variation in inference time based on different input lengths \\
    GPT 3.5 & 3-5 secs & CPU & Almost no variation in inference time based on different input lengths \\ 
    \end{tabular}
    \caption{Inference time comparison}
\end{table}

\subsubsection{Word Cloud Visualization}
Fig 4 shows the visualization of keywords extracted using Llama2-7B model for one of the PubMed abstracts. The keywords appearing larger in size matches the keywords extracted using Llama2-7B and keywords present in the index terms of an abstract from PubMed.

\begin{figure}[htbp]
\centerline{\includegraphics[width=110mm,height=8cm]{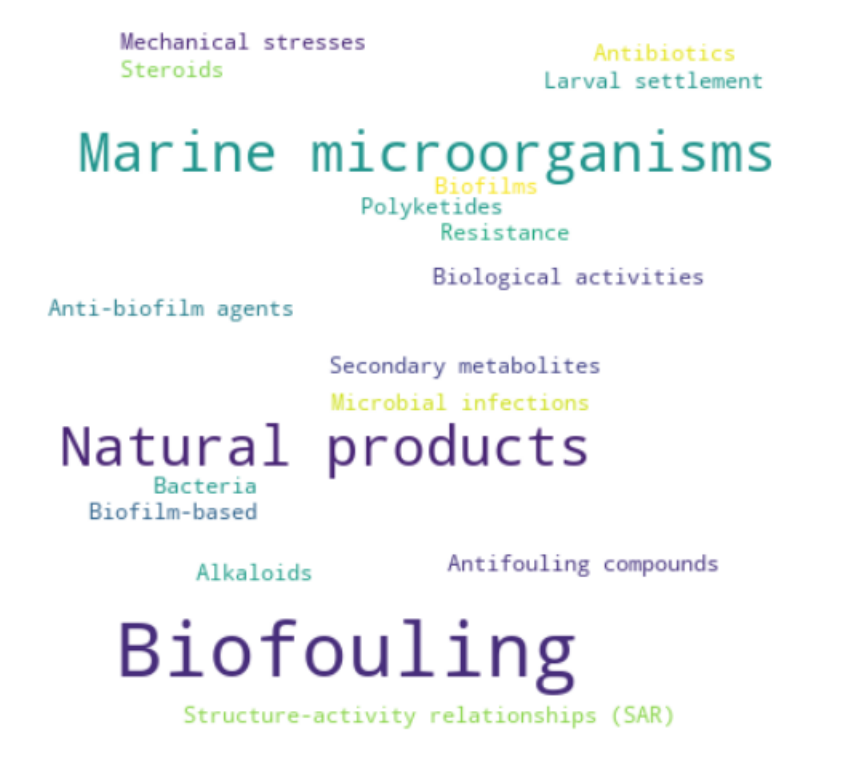}}
\caption{Keyword visualization with word clouds}
\label{wordcloud}
\end{figure}

% \section{Availability of package for Research and Development}
% To foster further research and collaborative development, the Python package is made publicly available. Interested parties can access the package, along with its documentation and source code, via [-------GitHub repository link------]. This repository includes comprehensive documentation, installation guides, and usage examples to assist researchers and developers in applying the package to their own keyword extraction tasks.

\section{Challenges and Future Implications}
There are several obstacles to overcome and long-term ramifications for keyword extraction utilizing large language models. Developing uniform evaluation criteria and benchmarks is necessary to address the variations among these models. The selection of suitable metrics introduces intricacy to the assessment procedure. Availability of high-quality reference datasets like Inspec and PubMed is essential, but ensuring their representativeness across domains remains a concern. Ethical considerations surrounding bias and potential misuse of generated keywords deserve attention. Looking ahead, we plan to expand the package's capabilities also. Future developments will focus on optimizing extraction algorithms, enhancing the package's scalability, and broadening its applicability to other NLP tasks. We also intend to explore the integration of additional LLMs and the incorporation of more advanced prompt engineering techniques. 

Large language models specifically developed for keyword extraction could be possible in future. Fine tuned models with domain specific dataset could enhance the keyword extraction techniques in near future. Understanding and mitigating the impact of hallucinations in model outputs is crucial for refining large language models and optimizing their performance in domain-specific tasks. Additionally, integrating human expertise and developing interactive, user friendly tools could enhance keyword extraction in various domains while maintaining ethical standards and addressing biases.
\section{Conclusion}
Large language models have brought about a tremendous revolution in the field of keyword extraction. In this review, we conducted a comprehensive exploration of keyword extraction techniques employing three prominent language models: Llama2-7B, Falcon-7B and GPT-3.5. Our exploration included the innovative application of prompt engineering techniques to guide these models more effectively. A key aspect of this research was the development and utilization of a specialized Python package, designed to streamline the keyword extraction process with these LLMs. This package, which seamlessly integrates with the LangChain framework, proved instrumental in efficiently interfacing with the models and extracting pertinent keywords.

Our goal was to assess the effectiveness of the LLMs in generating keywords and to compare their performance against reference Inspec and PubMed datasets. Llama2-7B exhibits a propensity to introduce additional terms, contributing to a broader scope, while Falcon-7B, despite presenting challenges with unnecessary words, demonstrates some competency in extracting pertinent keywords. GPT-3.5 emerges as a precision-focused model, generating concise key words with contextual alignment. We predict an exciting trajectory for keyword extraction skills as these language models advance. This review not only contributes to the current understanding of LLMs’ performance in keyword extraction but also provides valuable insights for researchers and practitioners. The ongoing evolution of LLMs in the context of keyword extraction opens up promising avenues for future research and applications, marking a transformative phase in the field of natural language processing.

% \bibliographystyle{plain}
% \bibliography{references}

% Here is a citation \cite{chow:68}.

% Acknowledgements and Disclosure of Funding should go at the end, before appendices and references

% \acks{All acknowledgements go at the end of the paper before appendices and references.
% Moreover, you are required to declare funding (financial activities supporting the
% submitted work) and competing interests (related financial activities outside the submitted work).
% More information about this disclosure can be found on the JMLR website.}


% \vskip 0.2in
% \bibliography{sample}

\end{document}